%% file: main.tex
\documentclass[10pt,twocolumn,letterpaper]{article}

\usepackage{iccv}              
\usepackage[whole]{bxcjkjatype}
\usepackage{multirow} 
\usepackage{graphicx}
\usepackage{amsmath}
\usepackage{subcaption}
\usepackage{amssymb}
\usepackage{booktabs}
\usepackage{epsfig}
\usepackage{multirow}
\usepackage{color}
\usepackage{newtxtext,newtxmath}
\usepackage{algorithm}
\usepackage[table]{xcolor} 
\usepackage{ulem}
\usepackage{xcolor}
\usepackage{colortbl}
\usepackage{url}

\definecolor{cvprblue}{rgb}{0.21,0.49,0.74}
\usepackage[pagebackref,breaklinks,colorlinks,citecolor=cvprblue]{hyperref}
\definecolor{Blue1}{rgb}{0.9,0.90,1} 
\definecolor{Blue2}{rgb}{0.6,0.84,1} 
\definecolor{Blue3}{rgb}{0.4,0.75,1} 

\newcommand{\first}[1]{\colorbox{Blue3}{#1}}
\newcommand{\second}[1]{\colorbox{Blue2}{#1}}
\newcommand{\third}[1]{\colorbox{Blue1}{#1}}

\def\paperID{15393} 
\def\confName{ICCV}
\def\confYear{2025}

\title{OD-RASE: Ontology-Driven Risk Assessment and Safety Enhancement for Autonomous Driving}
\author{$^{\dag}$Kota Shimomura$^{1,2}$, \quad $^{\dag}$Masaki Nambata$^{1,2}$, \quad Atsuya Ishikawa$^{3}$, \quad Ryota Mimura$^{3}$, \\
Koki Inoue$^{2}$, \quad Takayoshi Yamashita$^{1}$, \quad $^{\ddag}$Takayuki Kawabuchi$^{3}$\\
{\small $^{\dag}$Equal Contribution, \quad $^{\ddag}$Corresponding Author} \\
Chubu University$^{1}$, \quad Elith Inc.$^{2}$, \quad Honda R\&D Co., Ltd.$^{3}$\\
\url{https://kotashimomura.github.io/odrase/}
}

\begin{document}
\maketitle 
\input{sec/0_abstract}  
\input{sec/1_intro}

\input{sec/2_Related}
\input{sec/3_Methodology}
\input{sec/3.5_Models}

\input{sec/4_Experiments}
\input{sec/5_final}
{
    \small
    \bibliographystyle{ieeenat_fullname}
    \bibliography{main}
}

\input{sec/X_suppl}


\end{document}

%% file: sec/0_abstract.tex
\begin{abstract}
Although autonomous driving systems demonstrate high perception performance, they still face limitations when handling rare situations or complex road structures. 
Such road infrastructures are designed for human drivers, safety improvements are typically introduced only after accidents occur. 
This reactive approach poses a significant challenge for autonomous systems, which require proactive risk mitigation. 
To address this issue, we propose OD-RASE, a framework for enhancing the safety of autonomous driving systems by detecting road structures that cause traffic accidents and connecting these findings to infrastructure development. 
First, we formalize an ontology based on specialized domain knowledge of road traffic systems. 
In parallel, we generate infrastructure improvement proposals using a large-scale visual language model (LVLM) and use ontology-driven data filtering to enhance their reliability. 
This process automatically annotates improvement proposals on pre-accident road images, leading to the construction of a new dataset. 
Furthermore, we introduce the Baseline approach (OD-RASE model), which leverages LVLM and a diffusion model to produce both infrastructure improvement proposals and generated images of the improved road environment. 
Our experiments demonstrate that ontology-driven data filtering enables highly accurate prediction of accident-causing road structures and the corresponding improvement plans. 
We believe that this work contributes to the overall safety of traffic environments and marks an important step toward the broader adoption of autonomous driving systems.
\end{abstract}

\begin{figure}[ht] \centering \includegraphics[width=1.0\linewidth]{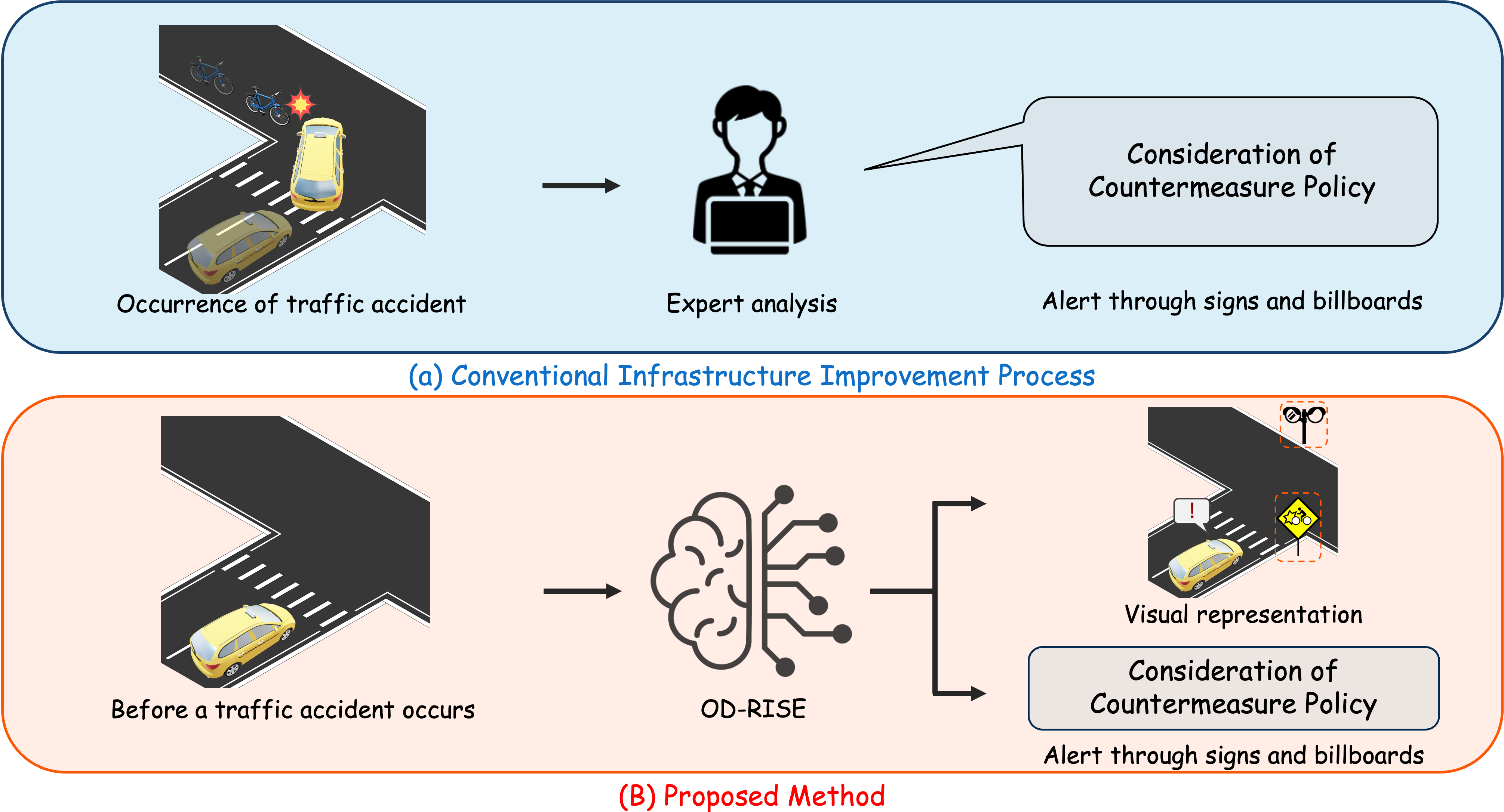} \caption{Comparison of various methods for infrastructure improvement design. (a) is based on expert knowledge, while (b) represents our proposed approach. Our method not only outputs infrastructure improvement plans for road structures that cause traffic accidents but also generates visual representations of roads after improvement.} \label{absfig} \end{figure}

%% file: sec/1_intro.tex
\section{Introduction}
State-of-the-art autonomous driving systems achieve highly accurate situational awareness by capturing information beyond what human drivers can process in real time, using approaches such as BEVPerception methods \cite{BEVFormer,tf,bevb} that efficiently extract 3D features from 2D images, as well as query-based multitask learning methods \cite{uniad,pd}. 
These efforts are supported by evaluation platforms that reproduce existing road structures to simulate arbitrary driving scenarios \cite{drivegs,EmerNeRF} and by benchmark datasets \cite{bdd,kitti,nuScenes,Cordts_2016_CVPR,mapi}. 
In addition, data-centric research \cite{li2024datacentric} has focused on generating arbitrary scenes \cite{chatsim,genad,pana} using diffusion models \cite{dfm} or novel view synthesis \cite{NeRF,3dgs} to address real-world corner cases. 
Furthermore, several datasets have been proposed to enhance safe autonomous driving by generating natural language reasoning explanations and evaluating systems ability to understand contextual information \cite{kim2018textual,9157111,8953562,chan2016anticipating}. 
Other methods explicitly learn regions from which vehicles or pedestrians might unexpectedly emerge during driving \cite{blinddetect,risk,occt}.

These studies substantially contribute to the safety of autonomous driving systems. 
Nevertheless, the design of road infrastructure has a critical impact on safety, including poorly visible intersections, lack of signage, insufficient sidewalk space, or sharp curves. 
Achieving a higher level of safety requires improving the transportation infrastructure that constitutes the driving environment. 
As shown in \cref{absfig}(a), conventional infrastructure improvements are typically carried out after a traffic accident has occurred, with experts analyzing the causes and proposing solutions based on their knowledge. 
When applied to autonomous driving systems, relying solely on reactive measures is insufficient for ensuring proactive safety in complex environments. It is essential to identify and mitigate potential environmental risks before they lead to critical situations.
Consequently, it is essential to expose potential risks in the road environment before an autonomous driving system causes an accident.

In this study, we propose a novel framework that enhances the safety of autonomous driving systems by preemptively identifying road structures that contribute to traffic accidents and connecting these insights to infrastructure improvements.
An overview of our framework is illustrated in \cref{absfig}(b). 

We first construct a dataset that is indispensable for generating infrastructure improvement proposals from images of road structures.
To that end, we leverage expert knowledge on road traffic systems to link road structures, their potential risks, and possible improvement methodologies.
We formalize this as an ontology based on expert knowledge. 
We then utilize a large-scale visual language model (LVLM) to generate infrastructure improvement proposals for each road structure. 
Next, we evaluate these proposals using our ontology, representing both the ontology and the proposals as graphs for graph matching.
Only proposals aligning closely with expert knowledge are selected to form our dataset. This ontology-driven filtering enhances the quality and reliability of the dataset. Subsequently, we train the OD-RASE model using the constructed dataset. 
The OD-RASE model comprises an image encoder \cite{res,vit,clip,long}, a text encoder \cite{roba,flan,long}, and a diffusion model. 
In addition to generating infrastructure improvement proposals for road structure images, the OD-RASE model can visualize the post-improvement road environment.
This capability enables various stakeholders, including urban planners and community members, to intuitively understand and discuss the benefits of potential improvements.

Our experiments on Mapillary Vistas \cite{mapi} and BDD100K \cite{bdd} demonstrate that our method can predict both accident-causing road structures and their corresponding improvement proposals. Moreover, our method shows high robustness in zero-shot prediction on regions outside the training data. These results indicate that our research offers a new perspective on increasing the safety of autonomous driving systems from the standpoint of improving road traffic environments. 

Our main contributions can be summarized as follows:
\begin{itemize} 
    \item We propose a novel framework that identifies road structures causing traffic accidents and links them to infrastructure improvements in advance.
    \item We formalize an ontology that leverages expert knowledge of road traffic systems to represent accident-causing road structures and corresponding infrastructure improvement proposals.
    \item We improve dataset quality and reliability through ontology-driven data filtering based on expert knowledge.
\end{itemize}

%% file: sec/2_Related.tex
\begin{figure*}[ht]
\centering
\includegraphics[width=1.0\linewidth]{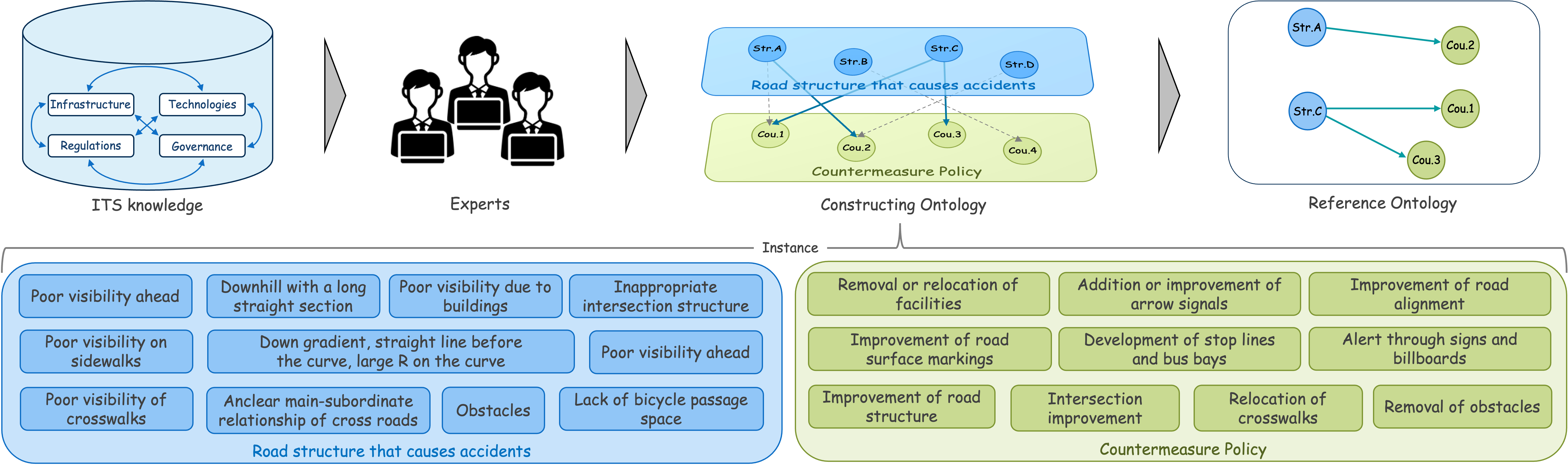}
\caption{Breakdown of infrastructure improvement process in field of road transportation systems, and overview of how OD-RASE Dataset is constructed on basis of it. Final set of 11 types of road structures causing traffic accidents (top) and 10 types of countermeasures (bottom).}
\label{ref}
\end{figure*}

\begin{figure}[ht]
\centering
\includegraphics[width=0.95\linewidth]{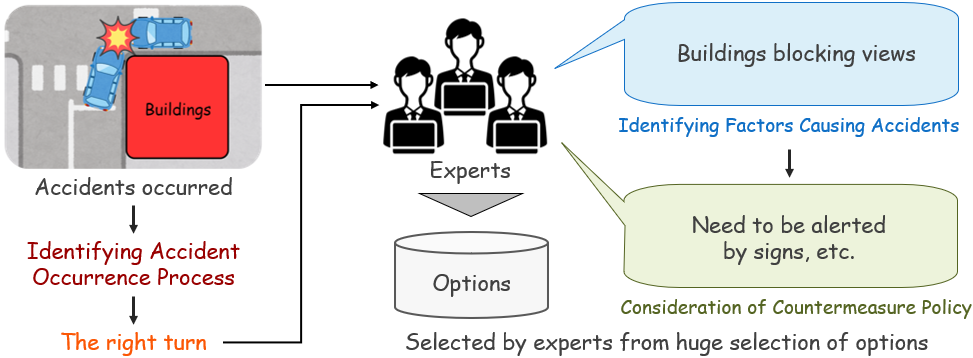}
\caption{Schematic diagram of the conventional  infrastructure improvement process by expert.}
\label{temp}
\end{figure}

\section{Related Work}
We first review research on improving road structures to prevent traffic accidents in the domain of road traffic systems. 
We then survey existing datasets aimed at mitigating traffic risks for autonomous driving.

\subsection{Road Infrastructure Improvements for Preventing Traffic Accidents}
An increasing number of initiatives seek to minimize the damage from accidents by improving road design under the assumption of driver-related human errors \cite{risk5,risk6}. 
In particular, modifying the road structure itself such as by reducing the number of lanes or installing medians is widely recognized as an effective measure for significantly decreasing the frequency of traffic accidents. 
In the United States, removing one or more lanes and adding turn lanes has achieved up to a 50\% reduction in accidents, and adding a continuous center turn lane can further reduce accidents by up to 65\% \cite{risk1,risk2}.

Converting intersections into roundabouts is another effective strategy \cite{risk2,risk3,risk4}. 
Roundabouts make it easier to control both the speed and angle of incoming vehicles and have been shown to reduce overall accident rates by around 38\%. 
Additionally, adding left  or right turn lanes, reshaping intersections, and installing pavement markings or warning signs have also proven effective \cite{risk8}. 
However, these infrastructure improvements primarily focus on human drivers rather than on enhancing the safety of autonomous driving systems. 
Consequently, there is a need to establish databases specifically designed to inform infrastructure improvements that align with the requirements of autonomous driving.

\subsection{Datasets for Autonomous Driving}
Datasets such as Anticipating Accident \cite{chan2016anticipating} and RAMS \cite{Ramanishka2018TowardDS}, along with those that provide textual explanations of risks in driving scenarios \cite{kim2018textual,9157111,8953562}, enable accident prediction and the identification or explanation of high-risk objects. 
More comprehensive approaches add risk annotations and natural language explanations to driving videos \cite{8953562,malla2023drama}. 
However, these datasets rely on manual annotations, making large-scale expansion challenging.

To automate the construction of natural language datasets, several approaches have used large-scale visual language models (LVLMs) or large language models (LLMs) for dataset generation \cite{chatsim,drivelm,speed}.
However, verifying the quality of generated annotations through human checks incurs a substantial overhead, so typically, only a small, randomly sampled subset undergoes manual review. 
Xie et al. point out that traditional LVLMs often generate plausible but potentially inaccurate answers by leveraging general knowledge or superficial textual cues \cite{syanhai}. 
Past work has primarily focused on predicting and describing objects or situations associated with traffic accidents and risks, neglecting the underlying road structures that can fundamentally cause these incidents. 
Additionally, while automated dataset construction methods reduce the burden of manual annotations, their reliability and quality checks are often limited to a small proportion of the generated data.

%% file: sec/3_Methodology.tex
\section{Ontology-Driven OD-RASE Dataset}
To train our OD-RASE model, we require a multimodal dataset consisting of images of road environments and corresponding infrastructure improvement proposals. 
However, no such dataset currently exists, necessitating its creation. 
Therefore, we attempt to build a dataset by leveraging the conventional process of infrastructure improvements.

\subsection{Structuring Infrastructure Improvement Process as Ontology}
\label{define}
As shown in \cref{temp}, conventional infrastructure improvement processes primarily focus on road environments where a traffic accident has already occurred. 
First, experts identify the conditions and triggering factors leading to the accident. 
Next, on the basis of their expert knowledge, they devise an infrastructure improvement process. 
This process can only be done post-accident. Additionally, because it relies on expert judgment, significant time may be required before an improvement proposal is finalized.

To ensure the safety of autonomous driving systems, it is essential to anticipate potential risks in road environments and perform improvements in advance, even if accidents have not yet occurred. 
In this research, we focus on summarized information on conventional infrastructure improvement processes~\cite{infra,risk9,risk10,risk11,risk12,risk13,risk14} and propose a method to structure these processes. 
This information includes more than 390 cases of road structures, accident conditions, and accident causes, but it lacks a structured format. 
Consequently, multiple experts in road traffic systems categorize road structures that can cause accidents into 30 types\footnote{See supplementary materials for details on the 30 types.}, assigning each structure a corresponding infrastructure improvement proposal.
Since some improvements overlap, we reduce the number of distinct proposals to 26 types\footnote{See supplementary materials for details on the 26 types.}. 
We then eliminate any time-dependent factors such as traffic volume or moving vehicles and consolidate similar elements into a single category.

Ultimately, as shown in ~\cref{ref}, the total set of accident-causing road structures is reduced to 11, with infrastructure improvement proposals narrowed down to 10.
The combinations of road structures and their respective improvement plans constitute the ontology derived from expert knowledge. 

\begin{figure}[ht]
\centering
\includegraphics[width=1.0\linewidth]{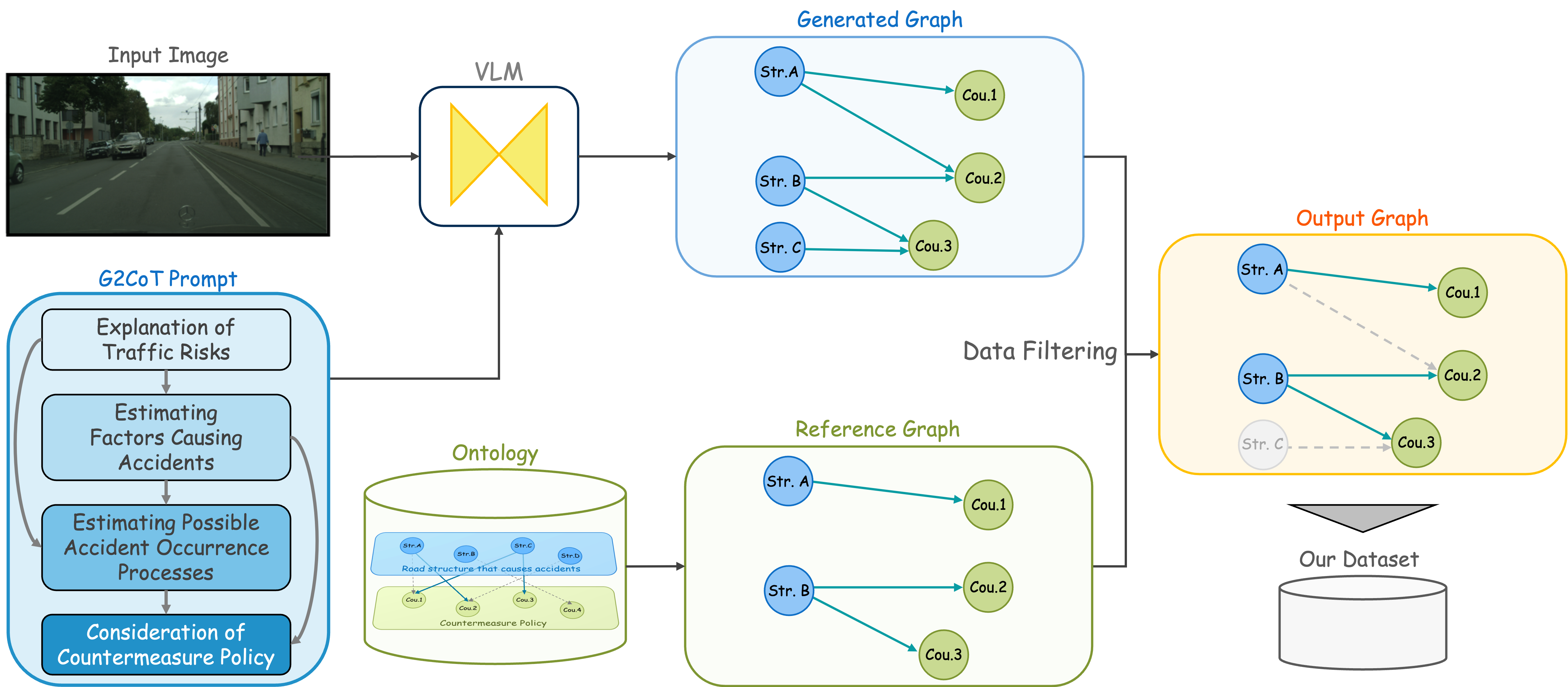}
\caption{Proposed ontology-driven dataset construction method. Our method allows for fully automatic generation using VLMs and enhances dataset quality and reliability through filtering with reference graphs. Additionally, such filtering further refines dataset's overall trustworthiness.}
\label{def}
\end{figure}
\subsection{Generating Infrastructure Improvement Proposals Based on Expert Reasoning}
\label{label}
~\cref{def} illustrates our method for generating candidate infrastructure improvements consistent with expert reasoning. 
To annotate any arbitrary image of a road structure with an infrastructure improvement plan, we emulate the way experts devise improvements. 
Experts first consider potential traffic-accident risks in a given image and then infer the conditions under which an accident could occur. 
Afterward, they predict the specific road structure that triggers the accident. 
This multi-stage inference process is provided to the VLM as a CoT (chain-of-thought) prompt~\cite{cot}. 
For the VLM, we adopt GPT-4o, which has shown high robustness in previous studies~\cite{syanhai} and delivers optimal performance on four autonomous driving tasks.

In this workflow, the text generated at each stage is converted into a graph-based prompt~\cite{drivelm} and passed to the subsequent stage. 
We refer to this process as the graph-based grounded CoT prompt (G2CoT)\footnote{For more information on this prompt, please refer to the supplementary materials.}. 
The infrastructure improvement proposals generated by G2CoT correspond to one of the 10 types described in ~\cref{define}, making it possible to convert them into ontology instances. 
Similarly, during the G2CoT prompt-generation step, the identified road structures also fall into one of the 11 types described in ~\cref{define}, enabling instance conversion. 
By leveraging these instantiations of road structures and infrastructure improvements to form our ontology, we can annotate any road-structure image with a proposed improvement plan. 

\subsection{Ontology-Driven Data Filtering Based on Expert Knowledge}
\label{dataset}
We evaluate the validity of the infrastructure improvement proposals generated by GPT-4o using the expert-knowledge ontology introduced in ~\cref{define}. 
\begin{figure}[ht]
\centering
\includegraphics[width=1.0\linewidth]{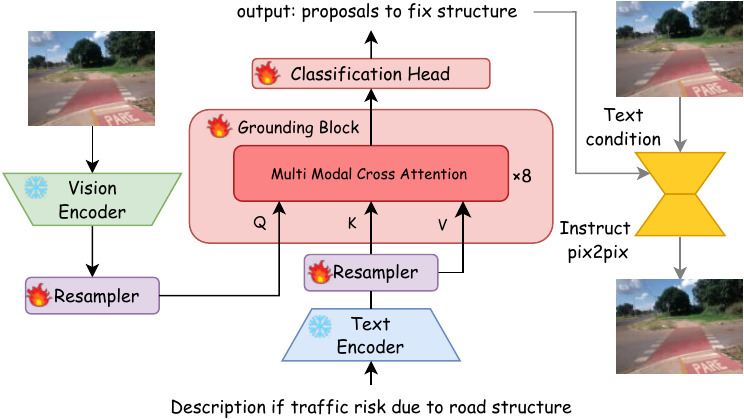}
\caption{OD-RASE model architecture. Each modality scene images and textual descriptions of traffic risks is encoded, and grounding block captures semantic relationships between image and text.}
\label{RISE}
\end{figure}
First, we take the ontology based on expert knowledge as a directed reference graph $G_A = (V_A, E_A)$.
We then treat the generated ontology of improvement proposals as a generated graph $G_B = (V_B, E_B)$. We evaluate these two graphs through graph matching.
To eliminate elements not present in the reference graph, we compute the common portions of the node sets and edge sets, as in ~\cref{ika}, creating a subgraph $G'_B = (V'_B, E'_B)$:
\begin{align} 
V'_B \;=\; V_B \,\cap\, V_A,  \quad E'_B \;=\; E_B \,\cap\, E_A.
\label{ika}
\end{align}

This step removes nodes ($V_B \setminus V_A$) and edges ($E_B \setminus E_A$) that conflict with expert knowledge. 
However, deleting nodes and edges in this manner may result in isolated nodes with no incoming or outgoing edges. 
In a directed graph $G' = (V', E')$, the set of isolated nodes is defined in ~\cref{iso}:
\begin{align} 
  \mathrm{Iso}(G') \;=\;
  \Bigl\{
    v \in V' \,\Big|\,
    \deg^-_{G'}(v) = 0 \;\wedge\;\deg^+_{G'}(v) = 0
  \Bigr\},
\label{iso}
\end{align}
where $\deg^-_{G'}(v)$ denotes the in-degree of node $v$ and $\deg^+_{G'}(v)$ denotes its out-degree.
To remove these isolated nodes, we apply ~\cref{de} to $G'_B = (V'_B, E'_B)$:
\begin{align} 
  V''_B \;=\; V'_B \,\setminus\, \mathrm{Iso}(G'_B).
\label{de}
\end{align}

We then retain only edges whose endpoints belong to $V''_B$, yielding the graph $G''_B$, which has been filtered in accordance with expert knowledge, as shown in ~\cref{las}:
\begin{align} 
  G''_B \;=\;
  \Bigl(
    V''_B,\;
    E'_B \,\cap\, (V''_B \times V''_B)
  \Bigr).
\label{las}
\end{align}

If all edges between any two modules are removed during filtering, we regard such data as untrustworthy and exclude it from the dataset. 
Through this process, all elements contradicting expert knowledge are discarded, and the final annotation data reflect the same line of reasoning as an expert, thereby achieving high quality.


%% file: sec/3.5_Models.tex
\section{OD-RASE Baseline}
The aim of this study is to enhance the safety of autonomous driving systems by highlighting potential risks in road structures that lead to traffic accidents.
To this end, we propose an architecture capable of predicting an improvement strategy for any road structure. 
As shown in ~\cref{RISE}, our proposed model not only forecasts infrastructure improvement plans but also uses a diffusion model to generate post-improvement road structures. 
This capability allows non-experts to visually assess the validity of the proposed solutions.

For novice engineers or non-experts, envisioning precise infrastructure improvements can be arduous. 
Introducing a diffusion model enhances interpretability, thereby facilitating communication among engineers and supporting decision-making.

\subsection{Multi-Modal Model for Long Contexts}
As shown in ~\cref{RISE}, our infrastructure improvement proposal model is composed of a vision encoder, a text encoder, and a grounding block.

\begin{table*}[t]
\small
\centering
\begin{tabular}{ll|cccc|cccc}
\toprule
\multirow{2}{*}{Vision Encoder} & \multirow{2}{*}{Text Encoder} & \multicolumn{4}{c}{Mapillary} & \multicolumn{4}{|c}{BDD100K}  \\
 & & Recall & Precision & F1 & Acc & Recall & Precision & F1 & Acc \\
\hline \hline
\multirow{3}{*}{ResNet-50\cite{res}} 
& RoBERTa-Base\cite{roba} 
& 58.21 & 73.54 & 64.98 & 37.12 & \cellcolor{Blue1}74.78 & 79.75 & 77.18 & 45.94 \\
& Flan-T5-xl\cite{flan} 
& \cellcolor{Blue2}64.96 & 14.42 & 23.60 & 0.00 & 72.85 & 17.06 & 27.64 & 0.00\\
& Long-CLIP\cite{long} 
& 57.98 & 75.54 & 65.60 & 37.98 & 72.69 & 80.96 & 76.60 & 45.68\\
\hline
\multirow{3}{*}{ViT-B\cite{vit}} 
& RoBERTa-Base\cite{roba} 
& 59.73 & \cellcolor{Blue3}78.35 & 67.79 & \cellcolor{Blue2}40.71 & \cellcolor{Blue3}74.94 & 81.62 & 78.14 & 47.98 \\
& Flan-T5-xl\cite{flan} 
& 57.26  & 9.81 & 16.74 & 0.00 & 47.70 & 8.74 & 14.78 & 0.00\\
& Long-CLIP\cite{long} 
& 59.57 & \cellcolor{Blue2}78.14 & 67.60 & 40.04 & 73.41 & \cellcolor{Blue3}83.31 & 78.05 & 47.79\\
\hline
\multirow{3}{*}{CLIP\cite{clip}} 
& RoBERTa-Base\cite{roba} 
& 63.70 & 77.09 & \cellcolor{Blue2}69.76 & \cellcolor{Blue2}40.71 & 74.68 & 82.29 & \cellcolor{Blue1}78.30 & \cellcolor{Blue2}48.69 \\
& Flan-T5-xl\cite{flan} 
& 9.96 & 3.20 & 4.85 & 0.00 & 7.42 & 2.61 & 3.86 & 0.00\\
& Long-CLIP\cite{long} 
& 60.47 & 77.10 & 67.78 & 39.96 & 73.07 & \cellcolor{Blue1}83.19 & 77.80 & 47.43 \\
\hline
\multirow{3}{*}{Long-CLIP\cite{long}} 
& RoBERTa-Base\cite{roba} 
& \cellcolor{Blue1}64.54 & 77.09 & \cellcolor{Blue3}70.26 & \cellcolor{Blue3}42.14 & \cellcolor{Blue2}74.83 & \cellcolor{Blue2}83.20 & \cellcolor{Blue3}78.79 & \cellcolor{Blue3}49.48 \\
& Flan-T5-xl\cite{flan} 
& \cellcolor{Blue3}65.67 & 14.98 & 24.39 & 0.00 & 73.05  & 17.49 & 28.22 & 0.00 \\
& Long-CLIP\cite{long} 
& 61.04 & \cellcolor{Blue1}77.21 & \cellcolor{Blue1}68.18 & \cellcolor{Blue1}40.21 & 74.65 & 82.85 & \cellcolor{Blue2}78.54 & \cellcolor{Blue1}48.62 \\
\bottomrule
\end{tabular}
\caption{Quantitative evaluation results for predicting infrastructure improvement proposals for road structures. The best second, third best performances are shown in \first{First}, \second{Second}, \third{Third}, respectively.}
\label{uselabel12}
\end{table*}

\noindent
\textbf{Text Encoder:}  
Let the batch size be $B$, the sequence length be $S$, and the embedding dimension be $d$. For an input text $\mathbf{x}^t$, we use a text encoder and obtain $E_t \in \mathbb{R}^{B \times S \times d}$, where $E_t$ is the latent representation of each text token. A linear projection is then used to produce the final text embedding $T \in \mathbb{R}^{B \times S \times d}$.

\noindent
\textbf{Vision Encoder:}  
Similarly, for an input image $\mathbf{x}^i$, we use a vision encoder to obtain $E^i \in \mathbb{R}^{B \times S' \times d}$, where $S'$ denotes the number of visual tokens. A linear projection is then used to produce the final image embedding $I \in \mathbb{R}^{B \times S \times d}$.

\noindent
\textbf{Grounding Block:}  
We treat the image embedding $I$ as the query and the text embedding $T$ as both the key and value, and we compute a multi-head cross attention as in ~\cref{eq:mhca}:
\begin{align}
\mathrm{CrossAttn}(\mathbf{Q}, \mathbf{K}, \mathbf{V})
= \mathrm{softmax}\!\Bigl(\frac{\mathbf{Q}\,\mathbf{K}^\mathsf{T}}{\sqrt{d_k}}\Bigr)\,\mathbf{V}.
\label{eq:mhca} 
\end{align}
The final attention output is averaged across the sequence dimension into a vector $c$. 
We then apply a fully connected layer to $c$ to predict the infrastructure improvement proposal. 
Since we have 10 predefined categories, the output is the class probability over these 10 categories.

\noindent
\textbf{Training:}  
An image of a road structure may have more than one valid improvement proposal. 
Thus, we formulate the task as multi-label classification. Given an input $x$ composed of both images and text, along with a target $y_c \in \{0,1\}$, the loss is defined as in ~\cref{eq:crossentropy}:
\begin{align}
\mathcal{L}
= - \sum_{c=1}^{C}
\Bigl[
  y_c \,\log p_c
  + 
  (1 - y_c) \,\log (1 - p_c)
\Bigr],
\label{eq:crossentropy}
\end{align}
where $C$ is the number of classes, and $p_c \in [0,1]$ is the class probability obtained by applying a sigmoid function to the model logits.

\subsection{Diffusion-Based Layout Control}
To visually depict post-improvement road structures, we adopt a diffusion model with Instruct Pix2Pix for layout control.
This module is intended to improve ease of use and explainability.
Although it does not directly contribute to the accuracy of forecasts, it plays an important role in the social implementation as a decision support tool.

The image editing process consists of two steps:

\begin{enumerate}
\item \textbf{Prompt Generation:} From the OD-RASE output, we create a text prompt describing the problematic elements of the road structure and the proposed improvements (e.g., \texttt{output1 and output2 and ...}). 
\item \textbf{Image Editing:} We feed both the original image and the generated text prompt into Instruct Pix2Pix~\cite{instp}, obtaining a new image with the road structure edited in accordance with the proposed improvements.
\end{enumerate}

Through this approach, we move beyond purely textual suggestions, offering a visually rendered depiction of the improved road structure.

%% file: sec/4_Experiments.tex
\section{Experiments}
In this section, we present experimental results aimed at addressing three questions: (1) Is it possible to derive infrastructure improvement proposals directly from road structure images? (2) Does utilizing an ontology based on expert knowledge enhance the quality of the dataset for infrastructure improvement proposals? (3) Can the proposed method predict infrastructure improvement plans for unseen road structures?

\noindent
\textbf{Training Details.}  
Below, we summarize our experimental setup. 
To enable learning of potential traffic accident risks across diverse regions and driving scenarios, we used the Mapillary Vistas \cite{mapi} and BDD100K \cite{bdd} datasets, but the proposed approach can be applied agnostically to other datasets. 
First, we generated candidate infrastructure improvement proposals for these datasets as described in ~\cref{label}.

\begin{table}[!t]
\centering
\small
\begin{tabular}{c|cccc}
\hline 
Modal & Precision & Recall & F1 & Acc \\
\hline 
Image & 57.42 & 72.37 & 64.03 & 34.50 \\
Text & 60.02 & 79.76 & 68.50 & 40.63 \\
\rowcolor{Blue3}
Image \& Text & 64.54 & 77.09 & 70.26 & 42.14 \\
\hline 
\end{tabular}
\caption{Ablation study on which modality (image, text, or both) is most effective for predicting infrastructure improvement proposals. }
\label{modalab}
\end{table}

Next, we evaluated the candidate proposals using the expert-knowledge ontology described in ~\cref{dataset}, resulting in final labels for infrastructure improvements. 
Note that for certain road structures, the final annotation may indicate no improvements needed. 
We used ResNet-50 \cite{res} and ViT-Base \cite{vit} as vision encoders and RoBERTa-Base \cite{roba}, Flan-T5-xl \cite{flan}, Long-CLIP \cite{long} as text encoders. 
Our OD-RASE model was then trained as a multi-label classification task with a batch size of 16 for 25 epochs. 
\begin{figure}[!t]
\centering
\includegraphics[width=0.95\linewidth]{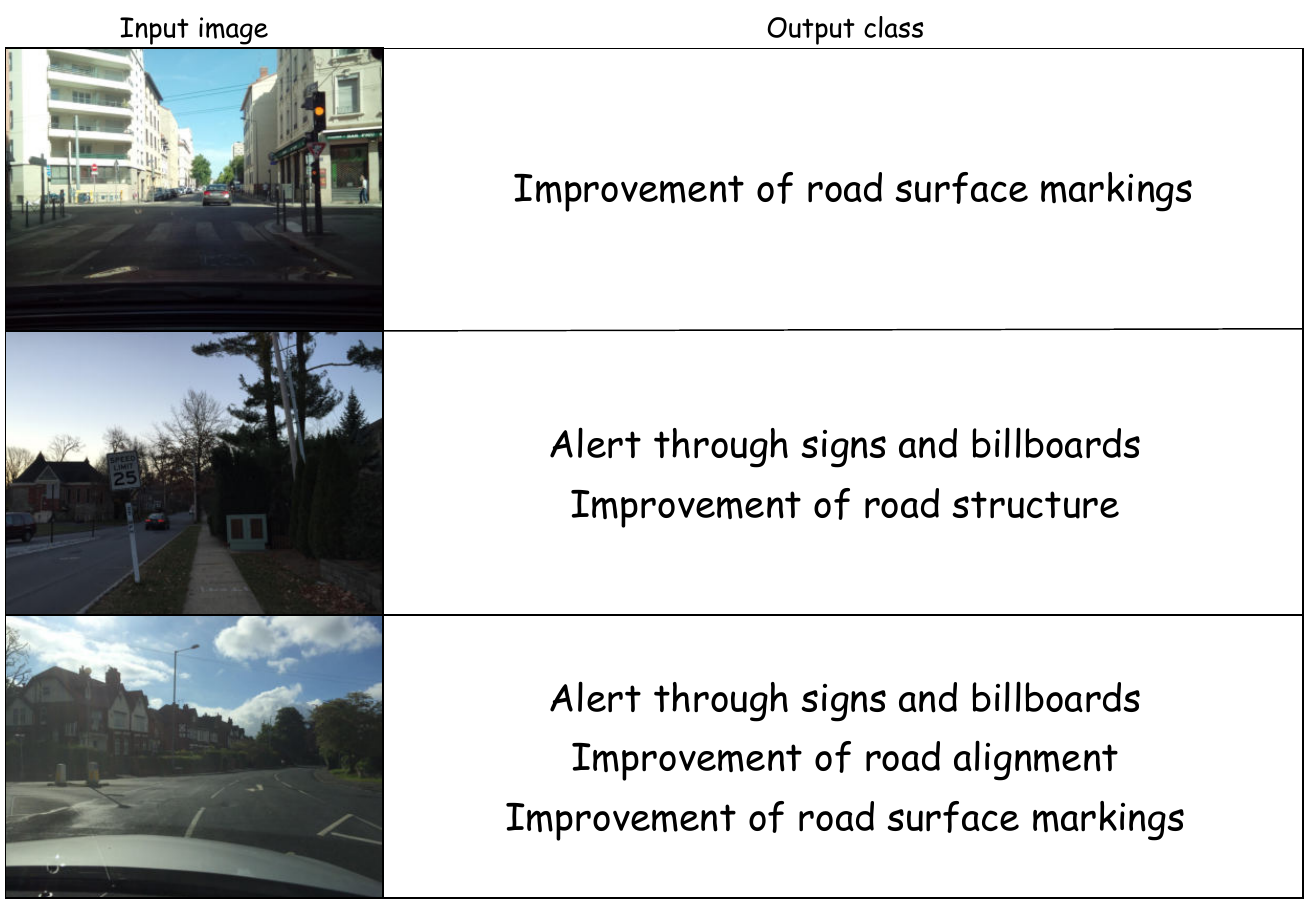}
\caption{Qualitative examples of infrastructure improvement proposals for road structures. Our OD-RASE model successfully predicted relevant infrastructure improvements.}
\label{ex1_fig}
\end{figure}
The parameters of both the vision and text encoders were frozen during training. 
We used Recall, Precision, F1-Score, and Accuracy, which are standard metrics for multi-label classification, for our evaluations.

\subsection{Predicting Infrastructure Improvements for Road Structures}
We first assessed the capability of our OD-RASE model to predict infrastructure improvement proposals for given road structures. 
According to ~\cref{uselabel12}, the model using Long-CLIP as the vision encoder and RoBERTa-Base as the text encoder achieved the highest performance on both the Mapillary and BDD100K datasets. 
\begin{table}[!t]
\centering
\small
\begin{tabular}{c|cccc}
\hline 
Data filtering & Precision & Recall & F1 & Acc \\
\hline 
 & 33.59 & 64.85 & 44.26 & 0.00 \\
\hline 
\rowcolor{Blue3}
\checkmark & 64.54 & 77.09 & 70.26 & 42.14 \\
\hline 
\end{tabular}
\caption{Ablation study on effectiveness of ontology-driven data filtering. Results indicate that filtering improved overall model performance.}
\label{abfil}
\end{table}
In contrast, when using Flan-T5-xl as the text encoder, we observed a high recall but low precision, suggesting an overestimation of candidate traffic risks and an increased rate of false positives. 
We conjecture that this is due to our use of 8-bit quantization on the pretrained weights.
~\cref{modalab} presents an ablation study focusing on different input modalities. 
The results show that combining images and text yielded the best accuracy when predicting infrastructure improvement proposals for road structures. 
Finally, we show qualitative examples in ~\cref{ex1_fig}.

\subsection{Effectiveness of Dataset Filtering}
We investigated the effectiveness of our ontology-driven dataset filtering method proposed in ~\cref{define}. 
In this experiment, we used Long-CLIP as the vision encoder and RoBERTa-Base as the text encoder, using Mapillary as our dataset.
\begin{table*}[!t]
\centering
\small
\begin{tabular}{l|l|cccccccc}
\toprule
\multicolumn{2}{c}{Method} & \multicolumn{2}{c}{Recall} & \multicolumn{2}{c}{Precision} & \multicolumn{2}{c}{F1-Score} & \multicolumn{2}{c}{Accuracy} \\
\hline
Vision Encoder & Text Encoder & val & test & val & test & val & test & val & test \\
\hline
\multicolumn{10}{c}{\textbf{Ours Baseline}} \\
\hline
ResNet-50\cite{res} & \multirow{4}{*}{RoBERTa-Base\cite{roba}} & \cellcolor{Blue3}64.38 & \cellcolor{Blue3}65.49 & 69.08 & 68.91 & \cellcolor{Blue1}66.65 & 67.16 & \cellcolor{Blue1}34.97 & 34.52 \\
ViT-B\cite{vit} & & \cellcolor{Blue1}62.49 & \cellcolor{Blue2}63.65 & \cellcolor{Blue2}74.34 & \cellcolor{Blue1}73.18 & \cellcolor{Blue2}67.90 & \cellcolor{Blue1}68.08 & \cellcolor{Blue2}38.02 & \cellcolor{Blue1}37.12 \\
CLIP\cite{clip} & & 60.64 & \cellcolor{Blue1}63.09 & \cellcolor{Blue1}73.76 & \cellcolor{Blue2}74.26 & 66.56 & \cellcolor{Blue2}68.22 & \cellcolor{Blue2}38.02 & \cellcolor{Blue2}37.28 \\
Long-CLIP\cite{long} & & \cellcolor{Blue2}62.68 & 62.34 & \cellcolor{Blue3}74.91 & \cellcolor{Blue3}75.57 & \cellcolor{Blue3}68.25 & \cellcolor{Blue3}68.32 & \cellcolor{Blue3}39.29 & \cellcolor{Blue3}38.96 \\
\hline
\multicolumn{10}{c}{\textbf{Generalist Models}} \\
\hline
\multicolumn{2}{c|}{GPT-4o\cite{gpt}} & 43.07 & 61.04 & 20.83 & 24.97 & 26.49 & 34.27 & 19.94 & 23.08 \\
\multicolumn{2}{c|}{LLaVA-1.5\cite{llava}} & 39.78 & 42.17 & 22.67 & 23.71 & 22.51 & 23.75 & 15.88 & 16.82 \\
\multicolumn{2}{c|}{Qwen2-VL\cite{Qwen}} & 41.68 & 40.99 & 17.94 & 17.93 & 24.09 & 23.97 & 15.46 & 15.55 \\
\multicolumn{2}{c|}{Phi-3\cite{phi}} & 41.70 & 41.27 & 22.01 & 22.13 & 27.48 & 27.47 & 17.47 & 17.48 \\
\multicolumn{2}{c|}{InternVL2\cite{inter}} & 38.23 & 34.57 & 14.87 & 14.18 & 20.17 & 18.83 & 12.73 & 11.97 \\
\bottomrule
\end{tabular}
\caption{Quantitative evaluation of our OD-RASE model versus generalist models on infrastructure improvement proposal tasks in zero-shot setting. Model was trained on BDD100K and evaluated on Mapillary.}
\label{zero1}
\end{table*}
\begin{table*}[!t]
\centering
\small
\begin{tabular}{l|l|cccccccc}
\toprule
\multicolumn{2}{c}{Method} & \multicolumn{2}{c}{Recall} & \multicolumn{2}{c}{Precision} & \multicolumn{2}{c}{F1-Score} & \multicolumn{2}{c}{Accuracy} \\
\hline
Vision Encoder & Text Encoder & val & test & val & test & val & test & val & test \\
\hline
\multicolumn{10}{c}{\textbf{Ours Baseline}} \\
\hline
ResNet-50\cite{res} & \multirow{4}{*}{RoBERTa-Base\cite{roba}} & 60.25 & 59.44 & 79.45 & 79.64 & 68.53 & 68.08 & 34.57 & 33.20 \\
ViT-B\cite{vit} & & \cellcolor{Blue1}63.00 & \cellcolor{Blue1}62.21 & \cellcolor{Blue3}85.46 & \cellcolor{Blue3}86.12 & \cellcolor{Blue1}72.53 & \cellcolor{Blue1}72.24 & \cellcolor{Blue1}41.00 & \cellcolor{Blue1}40.62 \\
CLIP\cite{clip} & & \cellcolor{Blue2}67.66 & \cellcolor{Blue2}66.86 & \cellcolor{Blue2}84.51 & \cellcolor{Blue2}84.87 & \cellcolor{Blue2}75.15 & \cellcolor{Blue2}74.80 & \cellcolor{Blue2}43.63 & \cellcolor{Blue2}43.17 \\
Long-CLIP\cite{long} & & \cellcolor{Blue3}70.22 & \cellcolor{Blue3}69.17 & \cellcolor{Blue1}83.68 & \cellcolor{Blue1}84.31 & \cellcolor{Blue3}76.36 & \cellcolor{Blue3}76.00 & \cellcolor{Blue3}45.79 & \cellcolor{Blue3}44.76 \\
\hline
\multicolumn{10}{c}{\textbf{Generalist Models}} \\
\hline
\multicolumn{2}{c|}{GPT-4o\cite{gpt}} & 54.26 & 50.73 & 22.79 & 21.54 & 31.19 & 29.38 & 21.64 & 20.40 \\
\multicolumn{2}{c|}{LLaVA-1.5\cite{llava}} & 47.21 & 46.71 & 26.39 & 26.91 & 27.27 & 27.47 & 19.50 & 19.62 \\
\multicolumn{2}{c|}{Qwen2-VL\cite{Qwen}} & 28.00 & 27.88 & 14.34 & 14.47 & 18.40 & 18.49 & 11.64 & 11.65 \\
\multicolumn{2}{c|}{Phi-3\cite{phi}} & 52.15 & 51.35 & 21.69 & 21.60 & 28.94 & 28.73 & 18.38 & 18.24 \\
\multicolumn{2}{c|}{InternVL2\cite{inter}} & 25.69 & 25.11 & 12.49 & 12.21 & 15.91 & 15.62 & 10.01 & 9.79 \\
\bottomrule
\end{tabular}
\caption{Quantitative evaluation of our OD-RASE model versus generalist models on infrastructure improvement proposal tasks in zero-shot setting. Model was trained on Mapillary and evaluated on BDD100K.}
\label{zero2}
\end{table*}
We trained models with and without expert-knowledge-based ontology filtering applied to the candidate proposals. 
The target task was to predict the road structure responsible for causing traffic accidents, and the evaluation was conducted on the post-filtered data.

~\cref{abfil} shows the performance with and without data filtering. 
When the training data were not filtered, the accuracy on the filtered evaluation set was notably low. 
Specifically, in the absence of filtering, the model demonstrated a high F1-Score of 44.26\,pt but a low Accuracy of 0.00\,pt, making it difficult to correctly identify the road structures leading to accidents. 
In contrast, once data filtering was used, the model achieved an F1-Score of 70.26\,pt and an Accuracy of 42.14\,pt, indicating a more robust learning outcome. 
From this experiment, we conclude that our proposed ontology-driven filtering, leveraging expert knowledge, is highly effective for exposing the potential risks of road structures that may trigger traffic accidents.

\subsection{Zero-Shot Prediction}
We next evaluated whether the proposed OD-RASE model can predict infrastructure improvement proposals for previously unseen road structures, using a zero-shot setting. 
We also examined whether state-of-the-art large-scale visual language models (LVLMs), referred to as generalist models, such as GPT-4o \cite{gpt}, LLaVA-1.5 \cite{llava}, and Qwen2-VL \cite{Qwen}, have any general knowledge of infrastructure improvements.

~\cref{zero1} shows the results for models trained on BDD100K and evaluated on the validation and test sets of Mapillary. 
Among our baseline variants, the combination of Long-CLIP as vision encoder and  RoBERTa-Base as text encoder yielded an F1-Score of 68.32\,pt and an Accuracy of 38.96\,pt on the Mapillary test set.
\begin{figure*}[!ht]
\centering
\includegraphics[width=0.95\linewidth]{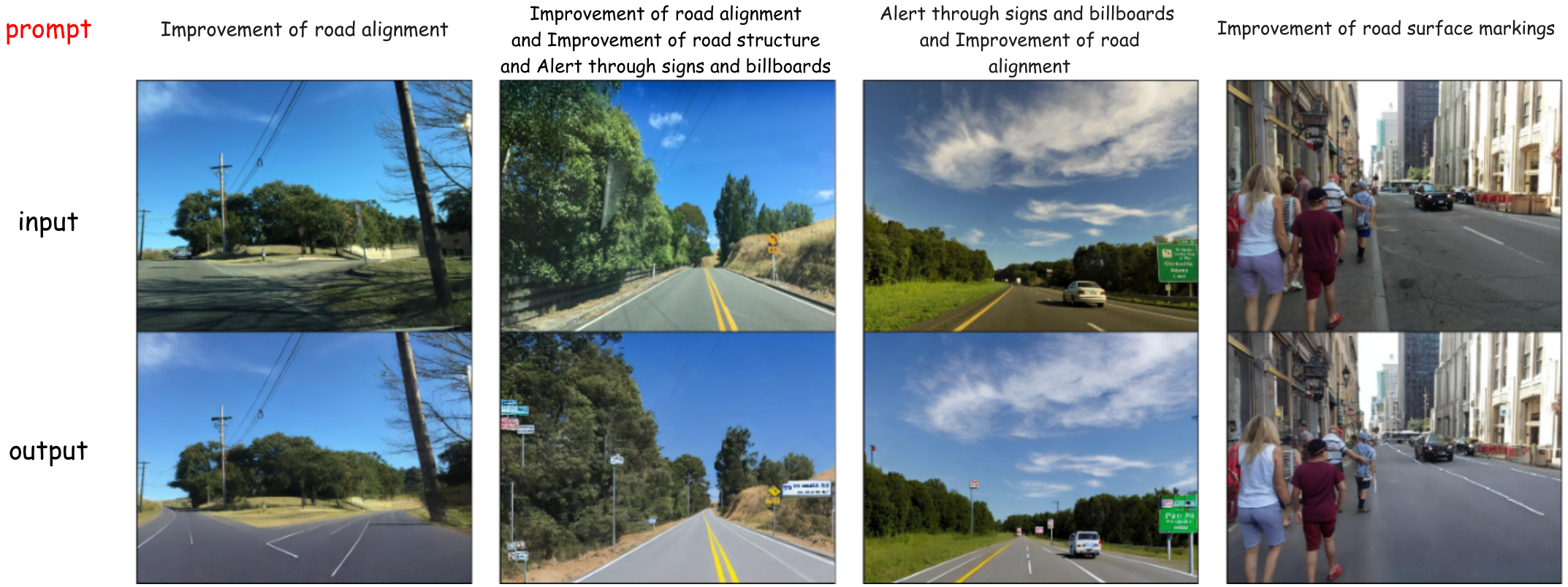}
\caption{Examples of layout control using \textit{Instruct Pix2Pix} \cite{instp}, based on OD-RASE-predicted infrastructure improvement plans.}
\label{diff}
\end{figure*}
In contrast, using a generalist model like Phi-3 yielded an F1-Score of 27.47\,pt and an Accuracy of 17.48\,pt, indicating difficulty in predicting improvement proposals for an unseen domain.

~\cref{zero2} shows the results for models trained on Mapillary and evaluated on the validation and test sets of BDD100K, confirming similar trends. 
These results highlight the necessity of expert-based ontology-driven data filtering. 

Indeed, our experiments on two distinct datasets and tasks confirm that existing generalist models lack sufficient domain knowledge to identify accident-causing road structures and propose suitable infrastructure improvements. 
Simply put, it is difficult to rely solely on these generalist models to detect and fix roads that may lead to traffic accidents.

\subsection{Diffusion-Based Layout Control} 
Finally, we show examples of images edited using Instruct Pix2Pix \cite{instp} based on the infrastructure development proposals predicted by the OD-RASE model in ~\cref{diff} and the quantitative evaluation results in ~\cref{ab1}.
As no ground-truth images for post-improvement roads were available, we provide a qualitative assessment and a quantitative assessment based on FID and expert human evaluation.
Experts judged each generated image as ``Full", ``Partial", or ``None" based on prompt Faithfulness.

~\cref{ab1} shows that the images generated by our method are highly interpretable and serve as an effective decision support tool for experts.
In the first column, where the proposal was ``Improvement to road alignment,'' the branching of the road was made more visible from a distance. 
In scenarios where the proposals were ``Improvement to road alignment, improvement to road structure, and alert through signs and billboards'' or ``Alert through signs and billboards and improvement to road alignment,'' the road changes to two lanes, and new signs were visibly added. 

Even when multiple proposals are combined, our method allows non-experts to intuitively understand the modifications through visual presentation. 
For ``Improvement of road surface markings,'' the white lane markers appear brighter and more pronounced, demonstrating the ability to generate images reflecting proposed improvements. 
Such visualizations can aid stakeholders in constructing road environments that enhance the safety of autonomous driving systems.

\begin{table}[t]
\centering
\begin{tabular}{c|c|ccc}
\hline 
\small
\multirow{2}{*}{Model} & \multirow{2}{*}{FID$\downarrow$} & \multicolumn{3}{c}{Prompt Faithfulness} \\
& & None & Partial  & Full\\
\hline 
Instruct Pix2Pix & 8.5 & 23.02 & 22.75 & 54.23 \\
\hline 
\end{tabular}
\caption{Quantitative evaluation results of the generated images.}
\label{ab1}
\end{table}

%% file: sec/5_final.tex
\section{Limitation}
Our current study has several limitations. 
We categorized road structure improvement proposals based on expert knowledge, excluding those requiring time-series analysis. 
Furthermore, our analysis relies solely on front-view images from onboard cameras. 
Future work could incorporate video inputs or comprehensive road-structure data (e.g., road gradients, intersection curvatures, GIS data) to suggest more sophisticated improvements.

Quantitatively assessing the effectiveness and accident reduction rates of our proposed improvements is challenging, as it would require a traffic simulator capable of editing road structures. 
Additionally, reflecting the importance and urgency of infrastructure improvement plans quantitatively remains difficult. 
However, the model's prediction probability can serve as a confidence level, and incorporating external data like regional traffic volumes and accident rates could help account for importance and urgency more explicitly.

\section{Conclusion}
Although various studies have consolidated the infrastructure improvement process into references \cite{infra,risk9,risk10,risk11,risk12,risk13,risk14}, these resources are neither structured nor applicable as datasets in computer vision contexts. 
Drawing on expert knowledge of road traffic systems, we structured road configurations, their potential risks, and methodologies for infrastructure development, defining them as an ontology. 
Using this ontology, we constructed a high-quality multimodal dataset of infrastructure improvement proposals. 
Experimental results show that by using ontology-driven data filtering, we can accurately predict both the road structures that cause accidents and their corresponding improvement plans. 
We also found that state-of-the-art LVLMs alone are insufficient for exposing such potential road risks. 
We believe our work paves the way for safer transportation environments and further advancement of autonomous driving systems, contributing not only to the safety of autonomous vehicles but also to the safety of all traffic participants, including pedestrians and manually driven vehicles.

%% file: sec/X_suppl.tex
\clearpage
\setcounter{page}{1}
\maketitlesupplementary

\section{Structuring Infrastructure Improvement Process as Ontology}
\label{sec:pruning_class}
The 30 types of accident-causing road structures and 26 types of infrastructure improvement proposals defined through expert knowledge include elements that overlap or are time-dependent (e.g., traffic volume, moving vehicles), which fall outside the scope of this research. 
Moreover, because these elements were derived from a wide variety of real-world traffic-accident cases, they are very granular and therefore not well-suited as a data structure for training our models.
For this reason, three experts reached consensus to merge similar elements and exclude those that are time-dependent.

\subsection{Consolidation of Similar Elements and Exclusion of Time-Dependent Elements}
\label{first}
The accident-causing road structures and countermeasure policies we defined are based on analyses of actual traffic accidents. 
Consequently, prior studies summarizing conventional infrastructure improvement processes \cite{infra,risk9,risk10,risk11,risk12,risk13,risk14} typically classify these elements at a fairly fine-grained level. 
As a result, multiple similar elements exist, which hinders effective model training.
Additionally, since the objective of this study is to analyze and improve risks arising from road infrastructure, any time-dependent factors must be removed. 
For these reasons, from the initially defined 30 accident-causing road structures and 26 infrastructure improvement proposals, we carried out the merging of similar elements and the exclusion of time-dependent elements. 
This step was performed by the same experts who created the dataset.

\begin{figure}[t]
\centering
\includegraphics[width=1.0\linewidth]{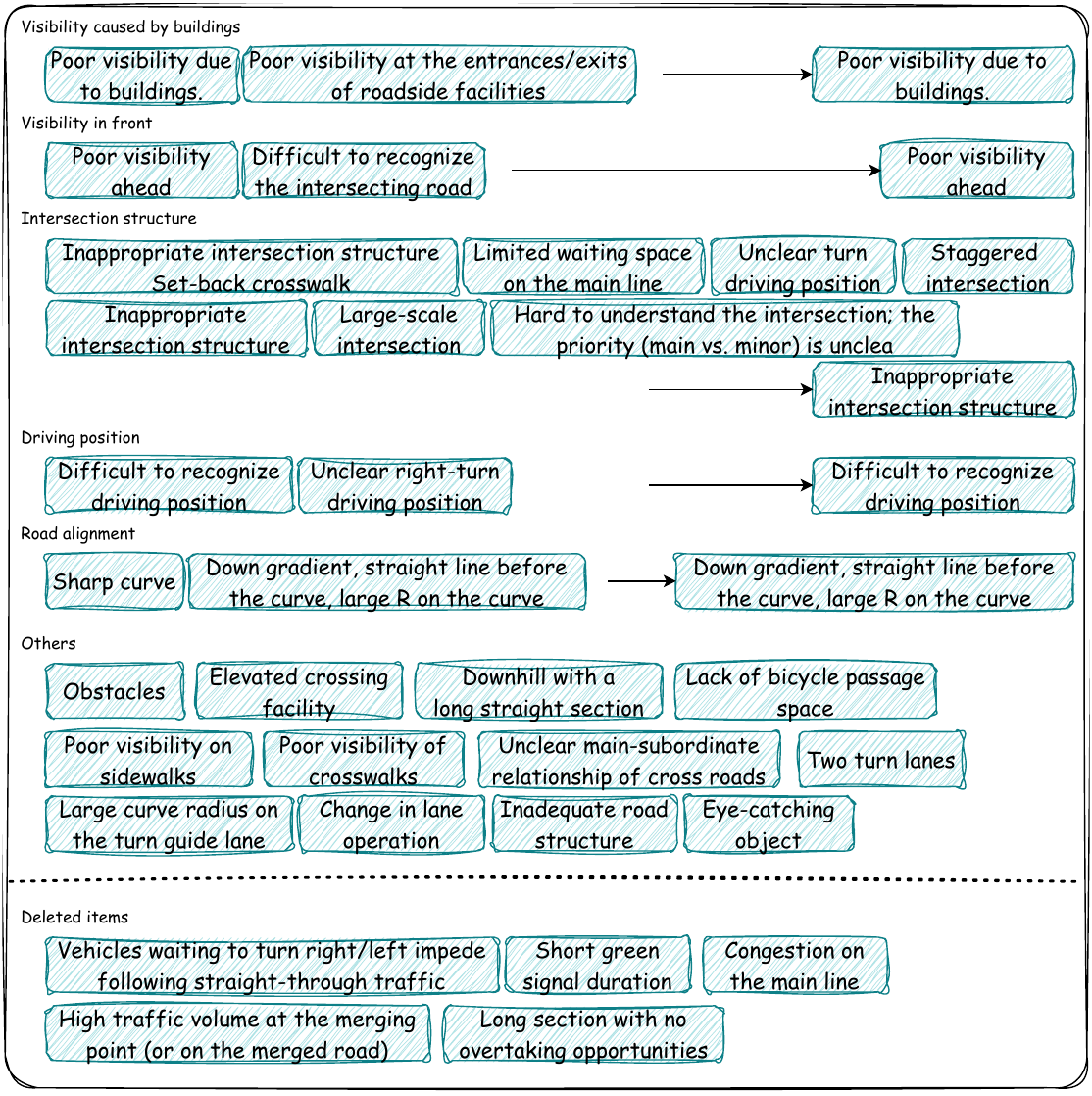}
\caption{Merging and exclusion of elements in accident-causing road structures.}
\label{fig:fca_step1}
\end{figure}

\begin{figure}[t]
\centering
\includegraphics[width=1.0\linewidth]{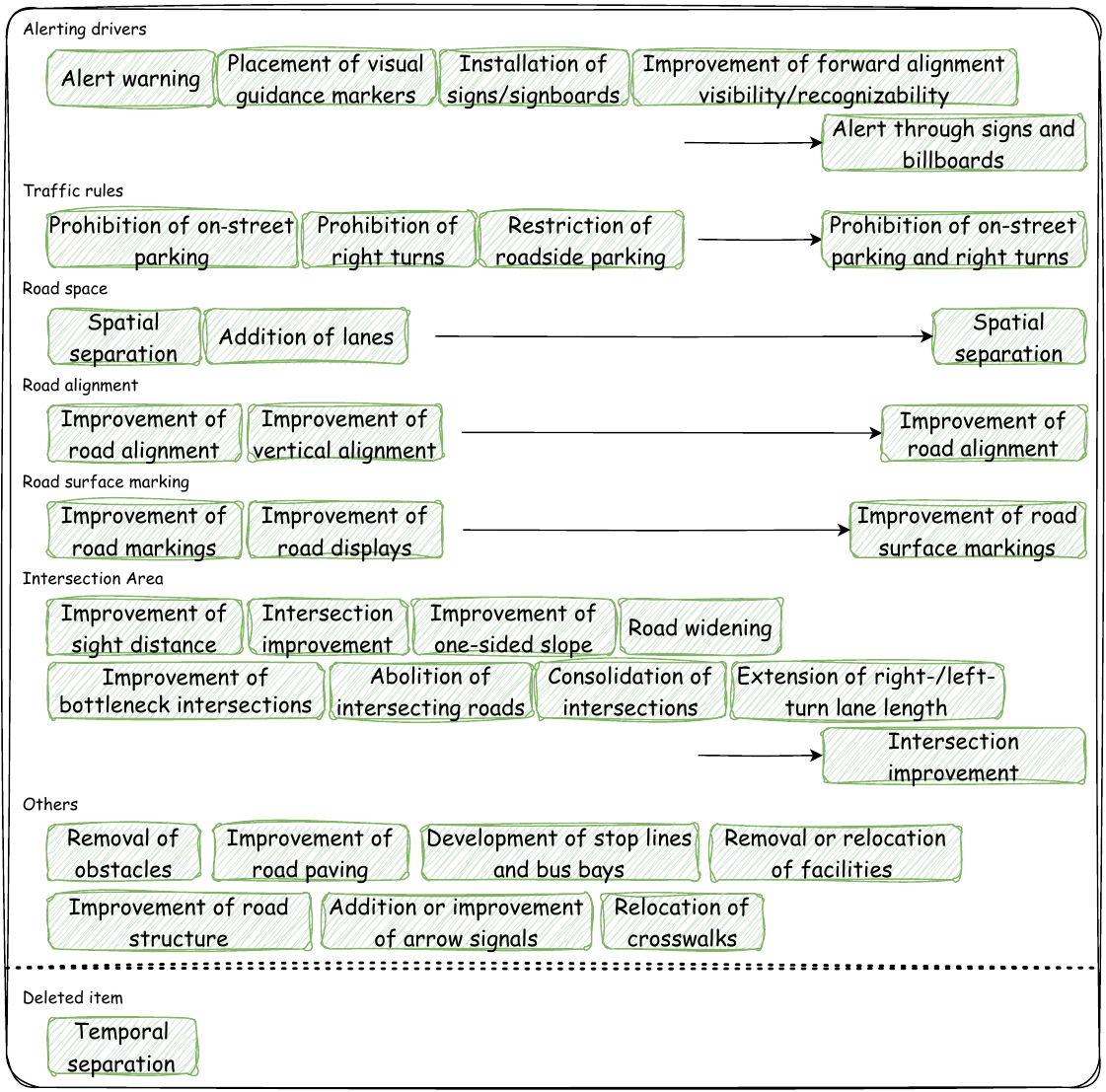}
\caption{Merging and exclusion of elements in infrastructure improvement proposals.}
\label{fig:cmp_step1}
\end{figure}

\cref{fig:fca_step1} and~\cref{fig:cmp_step1} illustrate the processes by which the accident-causing road structures and the infrastructure improvement proposals were merged or excluded, respectively. 
The merging of similar elements was grouped by road environment. As a result of these procedures, the accident-causing road structures were reduced to 15 types, and the infrastructure improvement proposals were reduced to 12 types.

\subsection{Exclusion of Elements Close to Corner Cases}
\begin{figure}[t]
\centering
\includegraphics[width=1.0\linewidth]{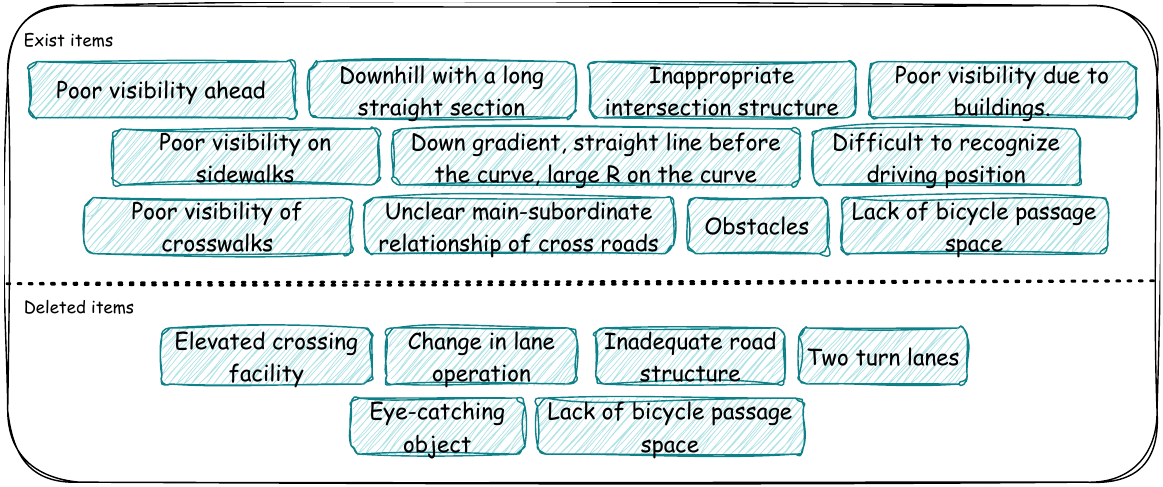}
\caption{Exclusion of accident-causing road structures related to corner cases.}
\label{fig:fca_step2}
\end{figure}

\begin{figure}[t]
\centering
\includegraphics[width=1.0\linewidth]{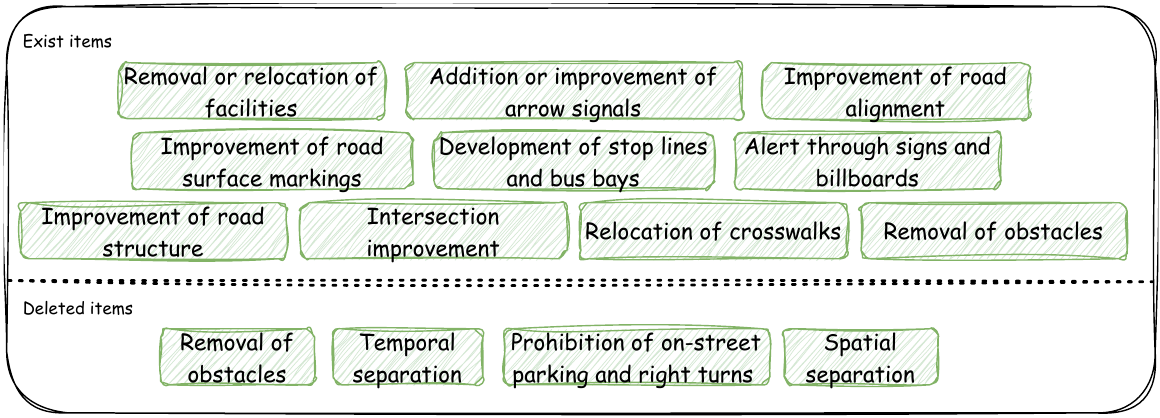}
\caption{Exclusion of corner-case elements in infrastructure improvement proposals.}
\label{fig:cmp_step2}
\end{figure}

In the \cref{first}, we consolidated elements by road environment and removed elements stemming from dynamic factors.
However, in real-world road environments, countless corner cases exist. 
Because our original definitions are quite granular, certain elements ended up disproportionately addressing corner-case situations.
In existing datasets as well, these corner-case elements appear so infrequently that they could risk further imbalancing the data.
Consequently, using expert knowledge, we carried out an additional exclusion of such corner-case elements from the accident-causing road structures and countermeasure policies that remained after the earlier merging and exclusion step.
An overview of this process is shown in \cref{fig:fca_step2} and \cref{fig:cmp_step2}. 
As a result, the final accident-causing road structures are reduced to 11 types and the infrastructure improvement proposals to 10 types.

\section{G2CoT:Graph-Based Grounded CoT Prompt}
\begin{figure*}[t]
\centering
\includegraphics[width=0.9\linewidth]{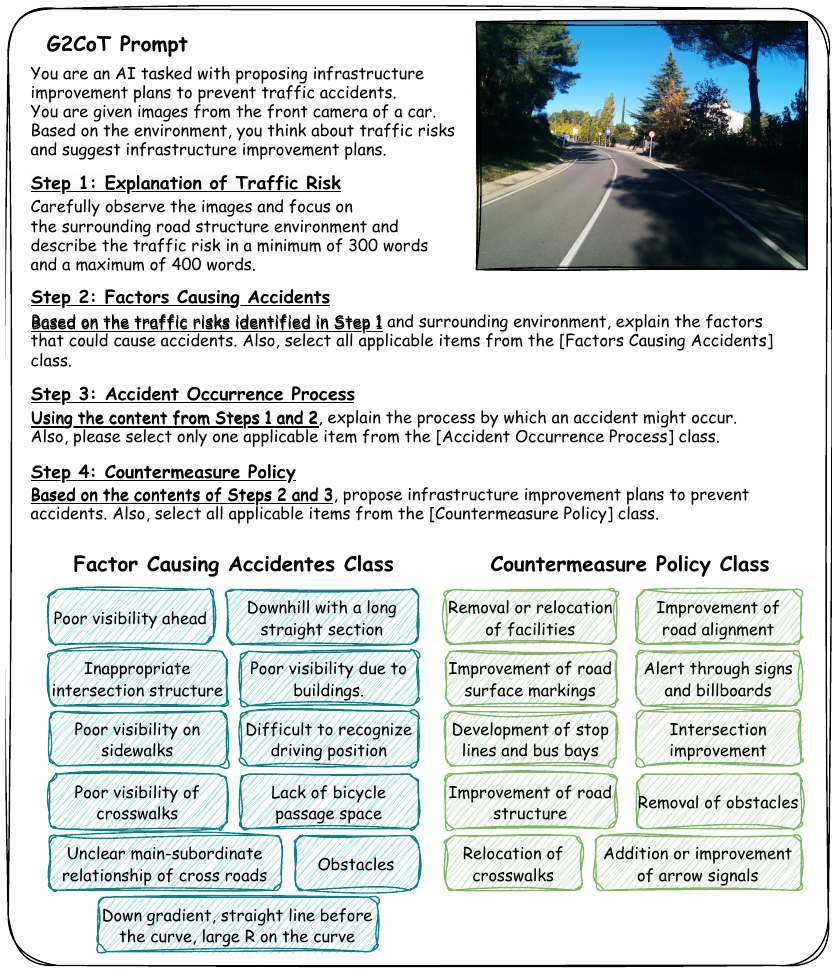}
\caption{Details of the G2CoT prompt used when constructing the OD-RASE Dataset.}
\label{fig:g2cot}
\end{figure*}

The proposed G2CoT uses a carefully designed CoT (chain-of-thought) prompt \cite{cot} to mimic the expert reasoning process when drafting infrastructure improvement proposals.
\cref{fig:g2cot} shows the details of our proposed G2CoT. 
As shown in \cref{fig:g2cot}, it generates outputs in four stages: (1) traffic risks, (2) accident-causing road structures, (3) accident occurrence processes, and (4) infrastructure improvement proposals. 
Specifically, Step 1 produces a textual explanation of static traffic risks from any given driving-scene image. 
In Step 2, referencing both the image and the results from Step 1, the model infers the accident-causing factors and selects all the elements that match from the accident-causing road structures we defined in \cref{sec:pruning_class}. 
In Step 3, referencing Steps 1 and 2, it predicts how an accident might unfold. 
Since infrastructure improvements for the same accident-causing factor can differ depending on the accident occurrence process, Step 3 aids in predicting infrastructure improvements by considering the accident process. 
Finally, in Step 4, referencing Steps 2 and 3, the model infers the infrastructure improvement proposals (i.e., countermeasure policies) and selects all applicable elements from those defined in \ref{sec:pruning_class}.
By performing this context-aware inference at each step, we can automatically build a dataset.

\begin{figure}[t]
\centering
\includegraphics[width=0.95\linewidth]{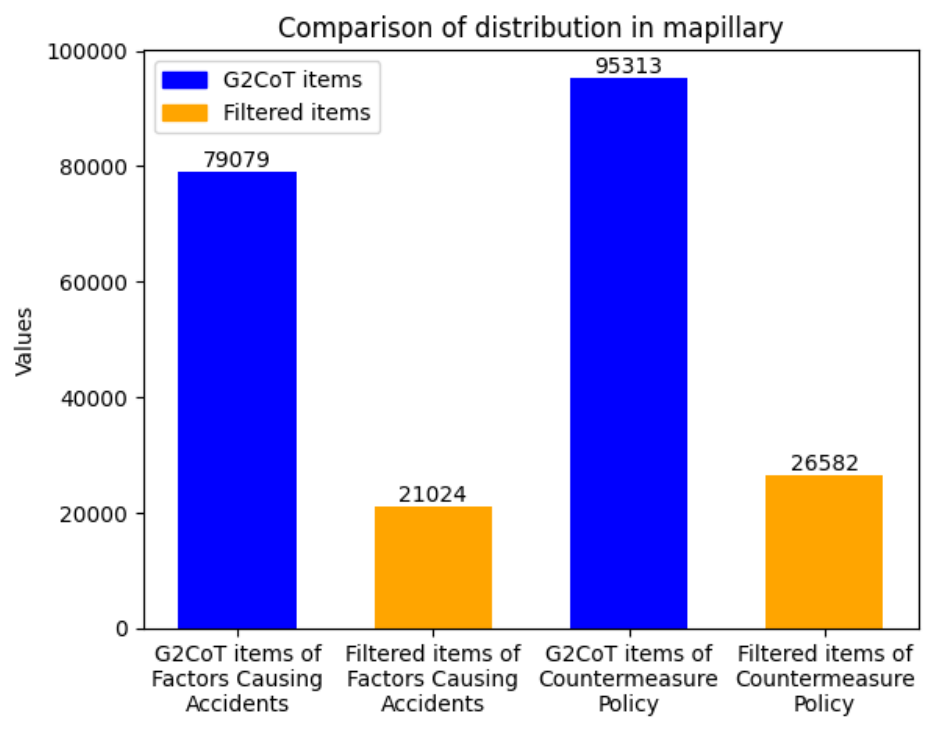}
\caption{Changes in data distribution before and after filtering for Mapillary Vistas. Blue: before filtering, orange: after filtering.}
\label{fig:mapi_dist}
\end{figure}

\begin{figure}[t]
\centering
\includegraphics[width=0.95\linewidth]{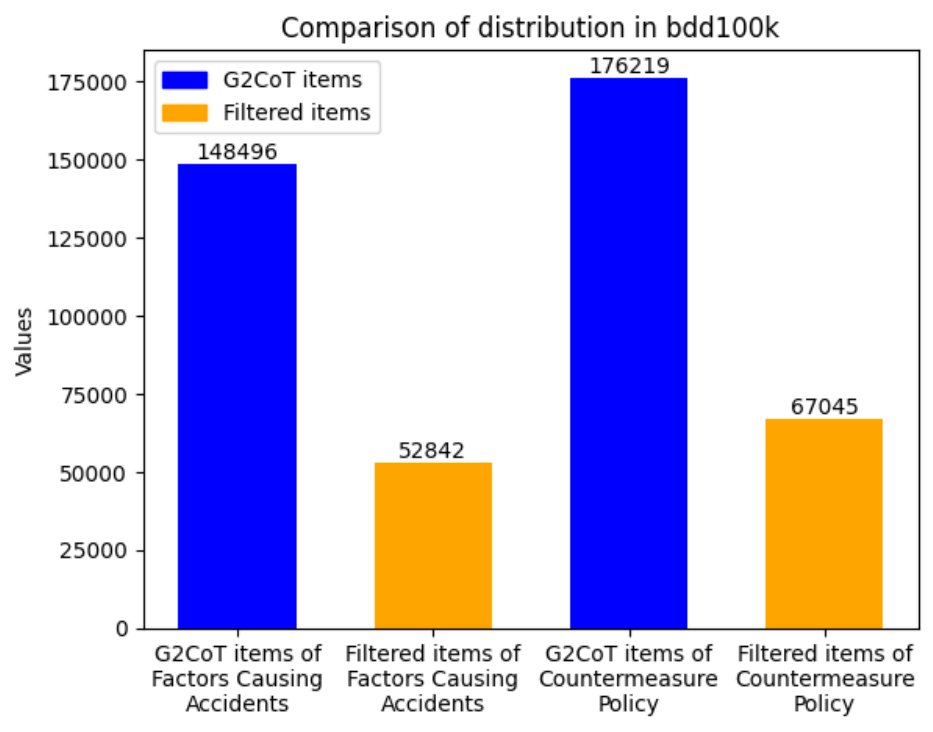}
\caption{Changes in data distribution before and after filtering for BDD100K. Blue: before filtering, orange: after filtering.}
\label{fig:bdd_dist}
\end{figure}

\begin{table*}[t]
\centering
\small
\begin{tabular}{ll|cccc|cccc}
\toprule
\multirow{2}{*}{Vision Encoder} & \multirow{2}{*}{Text Encoder} & \multicolumn{4}{c}{Mapillary} & \multicolumn{4}{|c}{BDD100K}  \\
 & & Recall & Precision & F1 & Acc & Recall & Precision & F1 & Acc \\
\hline \hline
\multirow{3}{*}{ResNet-50\cite{res}} & RoBERTa-Base\cite{roba} & 74.85 & \cellcolor{Blue1}82.06 & \cellcolor{Blue1}79.19 & \cellcolor{Blue1}64.25 & 86.54 & 88.55 & \cellcolor{Blue1}87.53 & \cellcolor{Blue1}73.50\\
 & Flan-T5-xl\cite{flan} & 72.07 & 12.50 & 21.30 & 0.00 & 73.94 & 13.54 & 22.89 & 0.00\\
 & Long-CLIP\cite{long} & \cellcolor{Blue3}77.17 & 76.96 & 77.06 & 58.92 & 83.35 & 88.74 & 85.96 & 71.04\\
\hline
\multirow{3}{*}{ViT-B\cite{vit}} & RoBERTa-Base\cite{roba} & 71.26 & 78.76 & 68.05 & 63.50 & \cellcolor{Blue1}88.80 & 86.44 & \cellcolor{Blue3}87.60 & 72.69 \\
 & Flan-T5-xl\cite{flan} & 29.01  & 3.76 & 6.65 & 0.00 & 24.25 & 3.33 & 5.85 & 0.00\\
 & Long-CLIP\cite{long} & \cellcolor{Blue1}76.64 & 80.53 & 78.54 & 61.68 & 85.27 & \cellcolor{Blue2}89.35 & 87.26 & 73.33\\
\hline
\multirow{3}{*}{CLIP\cite{clip}} & RoBERTa-Base\cite{roba} & 64.08 & 77.80 & 70.28 & 57.00 & \cellcolor{Blue3}89.17 & 85.92 & 87.51 & 72.36\\
 & Flan-T5-xl\cite{flan} & 24.21 & 4.44 & 7.50 & 0.00 & 22.58 & 4.28 & 7.20 & 0.00\\
 & Long-CLIP\cite{long} & \cellcolor{Blue2}77.10 & \cellcolor{Blue2}82.66 & \cellcolor{Blue2}79.79 & \cellcolor{Blue2}65.06 & 84.91 & \cellcolor{Blue3}90.02 & 87.39 & \cellcolor{Blue2}73.60 \\
\hline
\multirow{3}{*}{Long-CLIP\cite{long}} & RoBERTa-Base\cite{roba} & 64.08 & 77.80 & 70.28 & 57.00 & \cellcolor{Blue2}89.10 & 85.92 & 87.48 & 72.30 \\
 & Flan-T5-xl\cite{flan} & 27.80 & 6.50 & 10.54 & 0.00 & 24.07 & 5.74 & 9.27 & 0.00 \\
 & Long-CLIP\cite{long} & 76.44 & \cellcolor{Blue3}83.89 & \cellcolor{Blue3}79.99 & \cellcolor{Blue3}65.09 & 86.23 & \cellcolor{Blue1}{88.99} & \cellcolor{Blue2}87.59 & \cellcolor{Blue3}73.96 \\
\bottomrule
\end{tabular}
\caption{Quantitative evaluation results for predicting accident-causing road structures. The best, second and third best performances are shown in \first{First}, \second{Second}, \third{Third}, respectively.}
\label{tab:fca_sup}
\end{table*}

\subsection{Changes in Data Distribution}
We now provide a quantitative comparison of the data distribution before and after applying our expert-knowledge-based filtering on the dataset automatically constructed via G2CoT. 
We compare the distributions for accident-causing road structures and countermeasure policies. 
\cref{fig:mapi_dist} and \cref{fig:bdd_dist} show the changes in data distribution for Mapillary Vistas \cite{mapi} and BDD100K \cite{bdd}, respectively. 
From both figures, we see that the output from GPT-4o \cite{gpt} contains a substantial amount of incorrect data, resulting in more than 50\% of generated annotations being discarded by our data filtering.

\begin{table}[!h]
\centering
\small
\begin{tabular}{c|cccc}
\hline 
Modal & Precision & Recall & F1 & Acc \\
\hline 
CLIP & 54.98 & 72.54 & 62.55 & 32.60 \\
Long-CLIP & 66.34 & 57.94 & 61.86 & 28.22 \\
\rowcolor{Blue3}
Ours & 76.44 & 83.89 & 79.99 & 65.09 \\
\hline 
\end{tabular}
\caption{Ablation study on grounding block is most effective for predicting infrastructure improvement proposals. }
\label{abmodel}
\end{table}

\section{Versatility of the proposed method}
This research has demonstrated that our OD-RASE model can accurately predict infrastructure improvement proposals for road structures. We have also shown that it can generalize to unknown road structures. In this section, we show that OD-RASE is also capable of predicting accident-causing road structures, not just the infrastructure improvement proposals. We employ the Mapillary Vistas \cite{mapi} and BDD100K \cite{bdd} datasets.

\subsection{Predicting Accident-Causing Road Structures}
We evaluate the ability of OD-RASE to predict road structures that lead to traffic accidents, using supervised learning.
\cref{tab:fca_sup} presents quantitative evaluations on the Mapillary and BDD100K datasets. 
From \cref{tab:fca_sup}, it is evident that models using RoBERTa-Base~\cite{roba} or Long-CLIP~\cite{long} as text encoders successfully predict road structures that cause accidents, across multiple vision encoders. 
By contrast, models using Flan-T5-xl~\cite{flan} exhibit high recall but low precision, resulting in too many false positives. 
Overall, the best performance is obtained when both the vision and text encoders are Long-CLIP.

Table~\ref{abmodel} shows an ablation study evaluating the impact of the grounding block in OD-RASE.
For this experiment, we used Long-CLIP as the vision and text encoder and trained on Mapillary. 
\begin{table}[!h]
\centering
\small
\begin{tabular}{c|cccc}
\hline 
Data filtering & Precision & Recall & F1 & Acc \\
\hline 
 & 38.25 & 84.33 & 52.63 & 1.01 \\
\hline 
\rowcolor{Blue3}
\checkmark & 76.44 & 83.89 & 79.99 & 64.09 \\
\hline 
\end{tabular}
\caption{Ablation study on effectiveness of ontology-driven data filtering. Results indicate that filtering improved overall model performance.}
\label{fil}
\end{table}
From ~\cref{abmodel}, OD-RASE, which includes the grounding block that integrates image and text, outperforms vanilla CLIP or Long-CLIP by a large margin. 
This confirms that our proposed grounding block is effective.

\subsection{Effectiveness of Dataset Filtering}
In this experiment, we used Long-CLIP as the vision and text encoder, using Mapillary as our dataset.

~\cref{fil} shows the performance with and without data filtering. 
When the training data were not filtered, the accuracy on the filtered evaluation set was notably low. 
Specifically, in the absence of filtering, the model demonstrated a high F1-Score of 52.63\,pt but a low Accuracy of 1.01\,pt, making it difficult to correctly identify the road structures leading to accidents. 
In contrast, once data filtering was used, the model achieved an F1-Score of 79.99\,pt and an Accuracy of 64.09\,pt, indicating a more robust learning outcome.
These results strongly support the necessity of our ontology-based data filtering grounded in expert knowledge. 

\begin{table*}[t]
\centering
\small
\begin{tabular}{l|l|cccccccc}
\toprule
\multicolumn{2}{c}{Method} & \multicolumn{2}{c}{Recall} & \multicolumn{2}{c}{Precision} & \multicolumn{2}{c}{F1-Score} & \multicolumn{2}{c}{Accuracy} \\
\hline
Vision Encoder& Text Encoder & val & test & val & test & val & test & val & test \\
\hline 
\multicolumn{10}{c}{\textbf{Ours Baseline}} \\
\hline
ResNet-50\cite{res} & \multirow{4}{*}{Long-CLIP\cite{long}} & 71.68 & 74.76 & 77.28 & 79.76 & 74.38 & 77.18 & 58.48 & 61.38 \\
ViT-B\cite{vit} & & \cellcolor{Blue3}76.84 & \cellcolor{Blue3}77.77 & \cellcolor{Blue1}79.68 & \cellcolor{Blue1}80.58 & \cellcolor{Blue3}78.24 & \cellcolor{Blue2}79.15 & \cellcolor{Blue2}62.43 & \cellcolor{Blue2}63.83 \\
CLIP\cite{clip} & & \cellcolor{Blue1}74.75 & \cellcolor{Blue1}75.63 & \cellcolor{Blue3}80.81 & \cellcolor{Blue3}82.22 & \cellcolor{Blue1}77.66 & \cellcolor{Blue1}78.79 & \cellcolor{Blue1}61.68 & \cellcolor{Blue1}63.25 \\
Long-CLIP\cite{long} & & \cellcolor{Blue2}75.49 & \cellcolor{Blue2}77.10 & \cellcolor{Blue2}80.22 & \cellcolor{Blue2}81.54 & \cellcolor{Blue2}77.78 & \cellcolor{Blue3}79.26 & \cellcolor{Blue3}62.50 & \cellcolor{Blue3}64.11 \\
\hline 
\multicolumn{10}{c}{\textbf{Generalist Models}} \\
\hline
\multicolumn{2}{c|}{GPT-4o\cite{gpt}} & 47.19 & 51.87 & 17.86 & 19.66 & 25.27 & 27.81 & 16.86 & 18.59 \\
\multicolumn{2}{c|}{LLaVA-1.5\cite{llava}} & 75.52 & 75.74 & 14.71 & 15.51 & 22.03 & 22.93 & 14.05 & 14.86 \\
\multicolumn{2}{c|}{Qwen2-VL\cite{Qwen}} & 83.42 & 84.42 & 21.43 & 22.51 & 32.98 & 34.28 & 21.07 & 22.18 \\
\multicolumn{2}{c|}{Phi-3\cite{phi}} & 52.15 & 51.35 & 21.69 & 21.60 & 28.94 & 28.73 & 18.38 & 18.24 \\
\multicolumn{2}{c|}{InternVL2\cite{inter}} & 68.27 & 72.36 & 21.07 & 22.63 & 30.73 & 32.74 & 20.45 & 21.85 \\
\bottomrule
\end{tabular}
\caption{Zero-shot prediction of accident-causing road structures. Models are trained on BDD100K and evaluated on Mapillary.}
\label{tab:fca_zeroshot_sup2}
\end{table*}

\begin{table*}[t]
\centering
\small
\begin{tabular}{l|l|cccccccc}
\toprule
\multicolumn{2}{c}{Method} & \multicolumn{2}{c}{Recall} & \multicolumn{2}{c}{Precision} & \multicolumn{2}{c}{F1-Score} & \multicolumn{2}{c}{Accuracy} \\
\hline
Vision Encoder& Text Encoder & val & test & val & test & val & test & val & test \\
\hline 
\multicolumn{10}{c}{\textbf{Ours Baseline}} \\
\hline
ResNet-50\cite{res} & \multirow{4}{*}{Long-CLIP\cite{long}} & \cellcolor{Blue1}81.10 & \cellcolor{Blue1}81.50 & \cellcolor{Blue1}88.37 & \cellcolor{Blue1}88.16 & \cellcolor{Blue1}84.58 & \cellcolor{Blue1}84.70 & \cellcolor{Blue1}68.35 & \cellcolor{Blue1}68.24 \\
ViT-B\cite{vit} & & 75.12 & 76.64 & 82.14 & 83.74 & 78.47 & 80.03 & 63.02 & 65.00 \\
CLIP\cite{clip} & & \cellcolor{Blue2}81.70 & \cellcolor{Blue2}82.04 & \cellcolor{Blue3}90.86 & \cellcolor{Blue3}90.95 & \cellcolor{Blue2}86.04 & \cellcolor{Blue2}86.26 & \cellcolor{Blue2}71.15 & \cellcolor{Blue2}71.66 \\
Long-CLIP\cite{long} & & \cellcolor{Blue3}83.14 & \cellcolor{Blue3}83.72 & \cellcolor{Blue2}90.07 & \cellcolor{Blue2}89.96 & \cellcolor{Blue3}86.46 & \cellcolor{Blue3}86.73 & \cellcolor{Blue3}71.45 & \cellcolor{Blue3}72.02 \\
\hline 
\multicolumn{10}{c}{\textbf{Generalist Models}} \\
\hline
\multicolumn{2}{c|}{GPT-4o\cite{gpt}} & 51.31 & 48.32 & 19.92 & 18.66 & 27.94 & 26.21 & 18.88 & 17.71 \\
\multicolumn{2}{c|}{LLaVA-1.5\cite{llava}} & 73.88 & 75.60 & 17.34 & 18.16 & 25.00 & 26.21 & 16.30 & 17.08 \\
\multicolumn{2}{c|}{Qwen2-VL\cite{Qwen}} & 86.31 & 86.92 & 24.81 & 25.08 & 37.09 & 37.48 & 24.36 & 24.67 \\
\multicolumn{2}{c|}{Phi-3\cite{phi}} & 41.70 & 41.27 & 22.13 & 22.01 & 27.48 & 27.48 & 17.47 & 17.48 \\
\multicolumn{2}{c|}{InternVL2\cite{inter}} & 76.32 & 76.91 & 26.68 & 26.78 & 37.48 & 37.65 & 25.75 & 25.91 \\
\bottomrule
\end{tabular}
\caption{Zero-shot prediction of accident-causing road structures. Model was trained on Mapillary and evaluated on BDD100K.}
\label{tab:fca_zeroshot_sup}
\end{table*}

\subsection{Zero-shot Prediction}
We conduct zero-shot prediction experiments, analogous to the infrastructure improvement proposals task, for accident-causing road structures. 
~\cref{tab:fca_zeroshot_sup2} shows results for models trained on BDD100K and evaluated on Mapillary Vistas, and ~\cref{tab:fca_zeroshot_sup} for those trained on Mapillary Vistas and evaluated on BDD100K. 
In both cases, the vision and text encoder pair of Long-CLIP achieves the highest performance, for instance an F1-Score of 79.26~pt and Accuracy of 64.11\,pt in ~\cref{tab:fca_zeroshot_sup2}. 
In contrast, using a generalist model like Qwen2-VL~\cite{Qwen} yielded an F1-Score of 34.28\,pt and an Accuracy of 22.18\,pt, indicating that predicting the factors leading to accidents in unknown domains is difficult.

~\cref{tab:fca_zeroshot_sup} shows the results for models trained on Mapillary and evaluated on the validation and test sets of BDD100K. 
Among our baseline variants, the combination of Long-CLIP for both the vision encoder and text encoder yielded an F1-Score of 86.73\,pt and an Accuracy of 72.02\,pt on the Mapillary test set.
Our broad experiments show that generalist models alone struggle to identify road structures that accident-causing road structures or propose meaningful infrastructure improvements.